%% file: main.tex
\documentclass[journal]{IEEEtran}

\usepackage[utf8]{inputenc}
\usepackage[T1]{fontenc}
\usepackage[table,xcdraw]{xcolor}
\usepackage{graphicx} 
\usepackage{amsmath} 
\usepackage{xcolor} 
\usepackage{caption}
\usepackage{multirow} 
\usepackage{amsmath,amsfonts}
\usepackage{algorithmic}
\usepackage{algorithm}
\usepackage{array}
\usepackage{makecell}
\usepackage{booktabs}
\usepackage[caption=false,font=normalsize,labelfont=sf,textfont=sf]{subfig}
\usepackage{textcomp}
\usepackage{stfloats}
\usepackage{url}
\usepackage{verbatim}
\usepackage{graphicx}
\usepackage{cite}
\usepackage{amsmath}
\usepackage{amssymb}
\usepackage{pifont}
\usepackage{gensymb}

\definecolor{rblue}{rgb}{0,0.5,1}
\definecolor{awesome}{rgb}{1.0, 0.13, 0.32}
\definecolor{hollywoodcerise}{rgb}{0.96, 0.0, 0.63}
\definecolor{lasallegreen}{rgb}{0.03, 0.47, 0.19}
\definecolor{hanpurple}{rgb}{0.32, 0.09, 0.98}
\definecolor{green(pigment)}{rgb}{0.0, 0.65, 0.31}
\usepackage[pagebackref=false,breaklinks=true,colorlinks,bookmarks=false]{hyperref}
\hypersetup{colorlinks,linkcolor={red},citecolor={hanpurple},urlcolor={magenta}}  
\DeclareUnicodeCharacter{2061}{}
\begin{document}
\title{HGeo-TopoMap: Boosting Topological Mapping with Hierarchical Geometric Priors}

\author{Siyu Li$^{1,2}$, Kunyu Peng$^{3,\dag}$, Di Wen$^{3}$, Beiping Hou$^{1}$, Zhiyong Li$^{2}$, and Kailun Yang$^{2,\dag}$%
\thanks{This work was supported in part by the National Natural Science Foundation of China (Grant No. 62388101 and No. 62473139) and in part by the Hunan Provincial Research and Development Project (Grant No. 2025QK3019).}
\thanks{$^{1}$The authors are with the School of Automation and Electrical Engineering, Zhejiang University of Science and Technology, China.}
\thanks{$^{2}$The authors are with the School of Artificial Intelligence and Robotics and the National Engineering Research Center of Robot Visual Perception and Control Technology, Hunan University, China (email: kailun.yang@hnu.edu.cn).}
\thanks{$^{3}$The authors are with the Institute for Anthropomatics and Robotics, Karlsruhe Institute of Technology, Germany.}
\thanks{$^{\dag}$Corresponding authors: Kailun Yang and Kunyu Peng.}
}

\maketitle

\begin{abstract}
\input{contents/abstract}
\end{abstract}


\IEEEpeerreviewmaketitle

\section{Introduction}
\input{contents/introduction}

\section{Related Work}
\input{contents/related_work}

\section{Methodology}
\input{contents/method}

\section{Experiments}
\input{contents/experiments}
\section{Conclusion}
\input{contents/conclution}

\bibliographystyle{IEEEtran}
\bibliography{ref.bib}

\end{document}

%% file: contents/abstract.tex
Topological maps provide essential road structure and connectivity information for autonomous driving. However, accurate centerline modeling remains challenging due to the lack of explicit visual markings and the insufficient exploitation of geometric regularities in existing methods. To address these challenges, we propose HGeo-TopoMap, a hierarchical geometric prior learning framework that integrates road-structure guidance and geometric consistency into centerline instance decoding. Specifically, a Geometric Adaptive Learning (GAL) module introduces spatially adaptive cross-attention to incorporate road structure priors derived from inverse perspective mapping, where distance-aware attention biases guide instance queries toward spatially relevant structural information. Furthermore, we develop a hierarchical centerline decoding strategy that progressively predicts instance centers and their constituent points, facilitating the learning of geometric orientations and spatial configurations. Building upon this structured representation, a Geometric Consistency Learning (GCL) module imposes layer-wise contrastive supervision on matched instance queries using geometric groupings, exploiting geometric relationships among centerline instances to regularize instance representations and improve centerline detection. By jointly exploiting spatial structural guidance and instance-level geometric regularities, HGeo-TopoMap enhances centerline detection and topology reasoning. Extensive experiments on the OpenLane-V2 centerline and lane segment benchmarks demonstrate improved topological mapping accuracy. Meanwhile, the proposed method demonstrates stable performance under the robustness benchmark. The source code and model weights will be made publicly available at \url{https://github.com/lynn-yu/HGeo-TopoMap}.

%% file: contents/introduction.tex
Online mapping has gained increasing attention in autonomous driving due to its cost-effectiveness and adaptability~\cite{hdmapreview,mapqp,mo2026mapuq,zhang2025local}. Different map representations serve various downstream tasks. Rasterized maps represent road environments as spatial grids, whereas vectorized maps describe road geometry using points and polylines~\cite{hdmapnet,maptrv2}.
Beyond geometric information, topological maps characterize road centerlines and their connectivity, as well as their relationships with traffic signs~\cite{review}, providing essential information for downstream path planning~\cite{mappath,sun2026diffsemanticfusion}.

\input{figure/intro}

Topological mapping is typically performed in the Bird's-Eye-View (BEV) space, where road instances are detected and their topological relationships are inferred~\cite{toponet,topologic}. 
Existing methods have made considerable progress in this direction. TopoNet~\cite{toponet} adopts a DETR-based architecture to detect centerline and traffic sign instances while incorporating graph neural networks to model their topological relationships.
Observing that connected centerlines often share endpoints, TopoLogic~\cite{topologic} introduces endpoint distances as geometric priors to facilitate topology reasoning. LaneSegNet~\cite{lanesegnet} further exploits the structural correlation between lane boundaries and centerlines through auxiliary lane-related predictions. Despite these advances, accurate centerline detection remains challenging. Unlike road boundaries and lane markings, centerlines generally lack explicit visual markings, as illustrated in the upper panel of Fig.~\ref{fig:intro}. Consequently, relying primarily on visual features may lead to ambiguous centerline representations, particularly in complex driving environments. This motivates the exploration of additional geometric priors to guide centerline instance learning.

\input{figure/intro_1}

One promising source of geometric guidance lies in the surrounding road structures. Although centerlines are not directly observable, their spatial distributions are closely associated with road boundaries and drivable regions. Recent advances in semantic segmentation~\cite{mask2former,segformer} have enabled the extraction of informative road structures from perspective images. These structures can be projected into the BEV space through Inverse Perspective Mapping (IPM), providing an explicit road structure prior without requiring a learned view transformation~\cite{genmapping}.
However, the resulting road structure maps may contain inaccuracies caused by occlusions, segmentation errors, and projection distortions. Directly incorporating these imperfect priors may introduce irrelevant information into centerline representations. Therefore, it is desirable to adaptively exploit the spatial relationships between centerline queries and road structures, allowing each query to selectively aggregate relevant prior information.

In addition to their relationships with surrounding road structures, centerlines exhibit intrinsic geometric regularities. As shown in Fig.~\ref{fig:intro}, centerlines can be broadly categorized into straight and curved instances, corresponding to different road configurations. Straight centerlines further exhibit characteristic orientation patterns, such as approximately parallel arrangements. These geometric attributes provide useful cues for distinguishing centerline configurations. Nevertheless, conventional point-based instance decoding often predicts centerline points directly from reference points, without explicitly modeling their relative configurations around an instance center. Such a formulation may not fully exploit the geometric structure of centerline instances. Moreover, conventional detection objectives do not explicitly encourage centerline queries with similar geometric attributes to learn consistent representations. These observations motivate the integration of instance-centered geometric decoding and geometry-guided representation learning, enabling centerline queries to better capture intrinsic geometric regularities.

Motivated by these observations, we propose HGeo-TopoMap, a hierarchical geometric prior learning framework that incorporates complementary geometric priors into topological map construction. Specifically, we develop a Geometric Adaptive Learning (GAL) module to exploit road structure maps derived from perspective images through IPM. GAL encodes spatial and semantic information from the road structure map and introduces distance-aware cross-attention to adaptively aggregate structural priors according to their spatial relationships with instance queries. Furthermore, we propose a Geometric Consistency Learning (GCL) module that integrates hierarchical instance decoding with geometry-guided contrastive learning. Within each decoder layer, centerline instances are progressively represented by predicting instance centers from reference points and subsequently reconstructing centerline points relative to their corresponding centers. This instance-centered representation facilitates the modeling of geometric orientations and spatial configurations. Based on this structured decoding process, GCL imposes layer-wise contrastive supervision on instance queries using curvature- and orientation-based geometric groupings derived from matched ground-truth centerlines. This encourages geometrically similar instances to learn consistent representations, thereby enhancing geometry-aware centerline modeling. By jointly exploiting explicit road-structure guidance and implicit geometric regularities, HGeo-TopoMap improves centerline representation learning for detection and topology reasoning.

Extensive experiments on the OpenLane-V2 dataset~\cite{wang2023openlanev2}, including centerline and lane segment benchmarks, demonstrate the effectiveness of the proposed framework, as shown in Fig.~\ref{fig:intro_1}. The results show improvements in topological mapping accuracy across different baselines, together with enhanced robustness under challenging driving conditions.

In summary, our main contributions are as follows:
\begin{itemize}
    \item We propose HGeo-TopoMap, a hierarchical geometric prior learning framework that integrates explicit road structure maps and implicit geometric regularities to enhance topological map construction.

    \item We develop a geometric adaptive learning module that incorporates road structure priors through distance-aware cross-attention, enabling spatially adaptive aggregation of structural information for centerlines.

    \item We introduce a geometric consistency learning module that unifies hierarchical decoding and geometry-guided contrastive learning, exploiting the consistency of geometric attributes among centerline instances to regularize instance representations and improve detection accuracy.
\end{itemize}

%% file: figure/intro.tex
\begin{figure}[!t]
      \centering
      \includegraphics[scale=0.40]{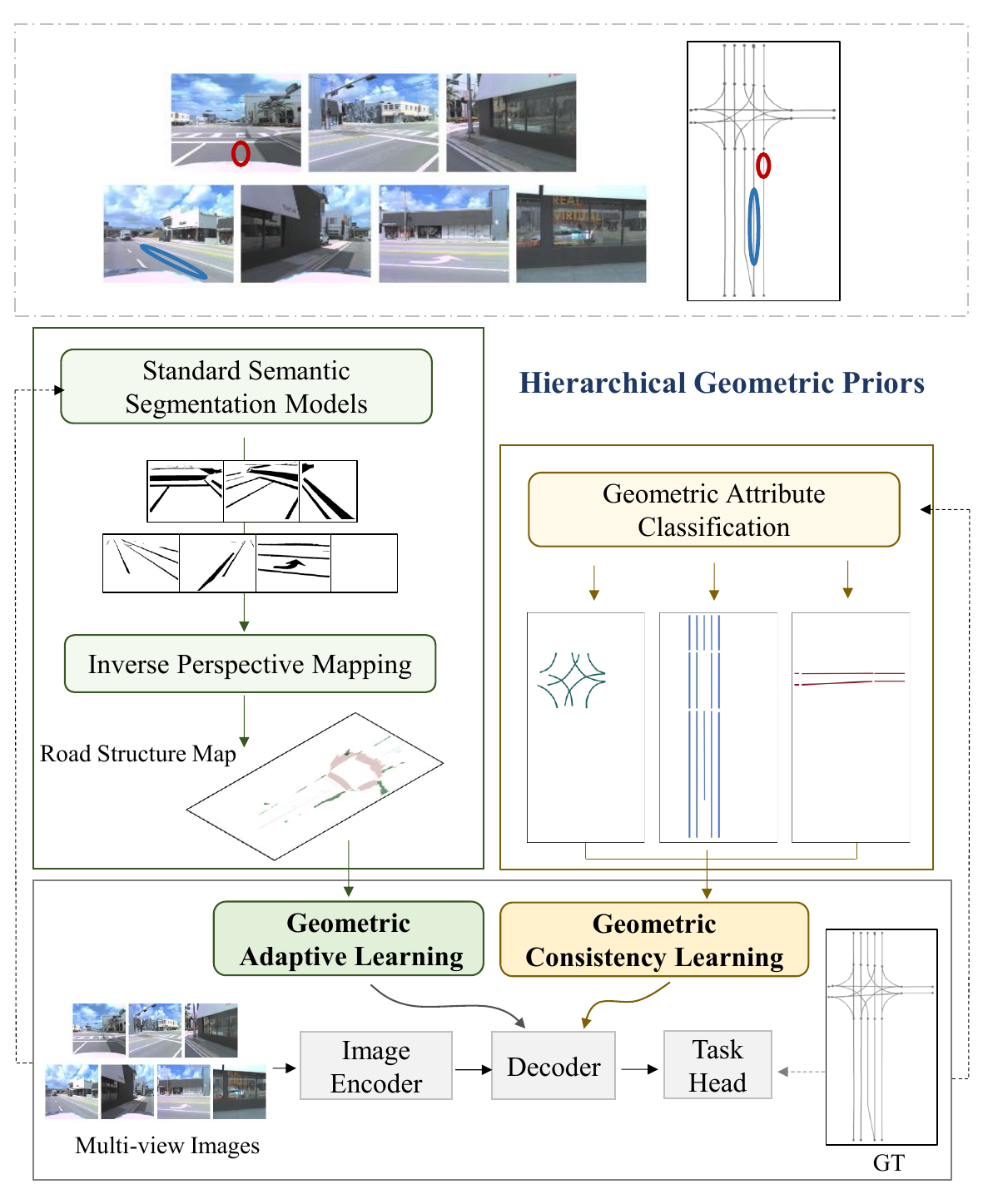}
      \caption{Motivation of HGeo-TopoMap. 
      To address the lack of explicit markings for instances, this paper introduces hierarchical geometric priors to guide topological mapping.
      }
      \label{fig:intro}  
\end{figure}

%% file: figure/intro_1.tex
\begin{figure}[!t]
      \centering
      \includegraphics[scale=0.42]{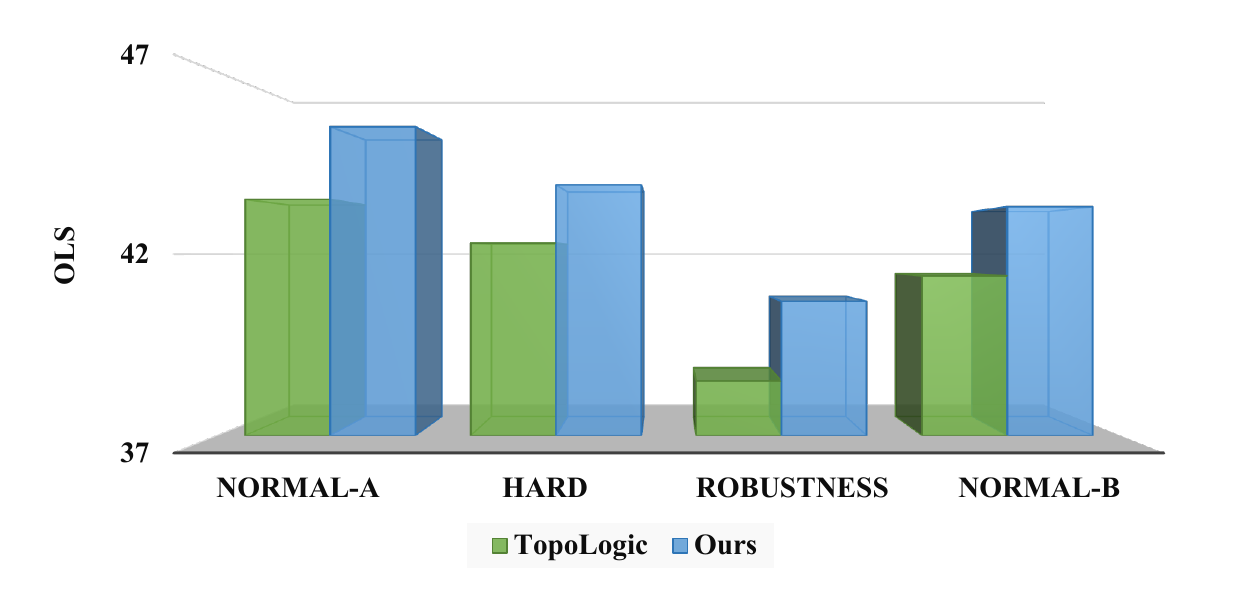}
      \caption{Comparison of results. The proposed method outperforms TopoLogic~\cite{topologic} under the normal split, the multi-lane hard split, robustness validation, and generalization validation.
      }
      \label{fig:intro_1}  
\end{figure}

%% file: contents/related_work.tex
Topological mapping is constructed in the Bird’s-Eye-View (BEV),  where the view transformer module from perspective views to the BEV serves as the core component.
Meanwhile, how to detect instances and learn topological relationships is also an important aspect.
Accordingly, this section describes, in order, view transformer techniques, topological mapping approaches, and priors for BEV mapping.

\subsection{View Transformer}
View transformer from perspective view to BEV is typically achieved through either implicit or explicit learning methods.
In map construction, explicit methods are more common. 
LSS~\cite{LSS} is a classic approach that introduced a depth estimation module to enable view transformer.
BEVDepth~\cite{bevdepth} extended this approach by incorporating LiDAR depth ground truth, thereby enhancing the accuracy of geometric prediction.
BEVFormer~\cite{li2022bevformer}, on the other hand, uniformly sampled in 3D space and leveraged a cross-attention mechanism to adaptively learn spatial features.
To improve computational efficiency, GKT~\cite{gkt} designed an index array composed of projection relationships.
In map construction tasks, since all road elements lie on the ground plane, their heights are known in real-world environments. Leveraging this prior knowledge, Inverse Perspective Mapping (IPM) achieves view transformation of road elements with strong generalization capability~\cite{PanopticBEV,genmapping}.
Therefore, this paper adopts IPM as the generation paradigm for the prior road structure map.

\subsection{Topological Mapping}

The topological mapping task entails the detection of traffic sign and centerline instances, followed by learning the connectivity relationships between them.
The detection of traffic signs commonly employs efficient backbones like Deformable DETR~\cite{zhu2020deformable} and YOLO~\cite{yolo}.
Normally, the detection process of centerline instances in topological mapping is highly similar to vectorized map tasks~\cite{liao2022maptr,Yuan_2024_streammapnet,mapqp}, employing a layer-wise regression learning approach.
Subsequently, researchers observed that connected centerline instances share endpoints. 
To improve topological map accuracy, they separately explored point-level and line-level detection~\cite{topo2seq,kalfaoglu2025topobda,topohr,li2025reusing}.
TopoPoint~\cite{fu2026topopoint} designed a dual detection module that jointly explores topological relationship learning at both point-level and line-level. 
Furthermore, UniTopo~\cite{unitopo} cast the learning of topological relations as new queries and employed multi-group learning to enhance efficiency.
TopoFG~\cite{xu2026fine_grained} refined query learning from the instance level to the point level and introduced a topological denoising training strategy to improve topological map accuracy.
To improve learning efficiency, the work~\cite{li2025reusing} proposed a one-stage decoder to learn traffic signs and centerline instances simultaneously.
After detecting traffic signs and centerline instances, graph convolutional networks~\cite{toponet} and multi-layer perceptrons~\cite{topomlp} are frequently employed as task heads to capture topological relationships between traffic signs and centerlines. 
These approaches perform topological reasoning on the basis of centerline instance detection, where topological accuracy hinges on detection performance. 
To enhance topology reasoning, RATopo~\cite{li2025ratopo} proposed a redundant assignment strategy that reorders the attention hierarchy, achieving a ``one-to-many'' assignment.

Simultaneously, several other approaches~\cite{topoformer,realtopo,kalfaoglu2026topomaskv3,liang2026persistent} exploited traffic knowledge to enhance topological learning performance.
TopoLogic~\cite{topologic} integrated geometric distances of centerlines into topology learning.
However, these works rely solely on distance relations among centerlines as prior guidance, while overlooking other geometric spatial relationships such as parallelism and perpendicularity.
To fully exploit these geometric relations, we introduce a consistency learning module for boosting topological mapping.

\subsection{Priors for BEV Mapping}
Generating maps from visual information faces the inherent sparsity problem of 2D data.
Consequently, embedding various priors has become a crucial approach to enhancing map quality.
P-MapNet~\cite{pmapnet}, SMART~\cite{SMART}, Score~\cite{pham2025coherent}, and NavMapFusion~\cite{monninger2026navmapfusion} used Standard Definition (SD) maps as priors for map construction.
Due to the misalignment in SD maps, P-MapNet employed standard cross-attention to enhance robustness.
Several other studies employ offline or historical High-Definition (HD) map data as priors~\cite{priorglobal,globalmapnet}.
Temporal cues provide valuable priors that mitigate the issue of sparse visual information~\cite{Yuan_2024_streammapnet,li2026amap}.
Moreover, D2HDMap~\cite{shon2026_nonvisible} and AerialFusionMapNet~\cite{lengerer2026aerialfusionmapnet} leveraged, respectively, driving paths and aerial images as auxiliary information to guide HD map construction.
Recently, road structure text has been introduced as a prior for map construction. SDTagNet~\cite{immel2025sdtagnet} encoded semantic information as text annotations, and TopoAware-Bench~\cite{chen2026vlms} built related visual question answering tasks, together capturing the core elements of spatial topological reasoning.
DiHMap~\cite{zhang2026dihmap} harnessed the priors of directional consistency to constrain geometric smoothness at the supervisory level.
MGMap~\cite{mgmap} and PercMap~\cite{priorvecmap} empowered vectorized map decoding with perspective road semantics and geometry, boosting both generalization and accuracy.
Similarly, Topo2D~\cite{topo2d} also employed 2D road structure information to assist 3D lane line detection.
Moreover, LaneSegNet~\cite{lanesegnet} used road structure information in an auxiliary task to boost centerline detection accuracy.
Nevertheless, these methods still require extra learning. 
Given the maturity of image-based road semantic segmentation, we argue that directly exploiting these priors is a more efficient and natural choice.
Accordingly, this work leverages the perspective road map as input and proposes a geometric adaptive module for centerline detection.

%% file: contents/method.tex
\subsection{Problem Formulation}

\subsubsection{Perspective Road Maps}
Map structures, such as dividers and pedestrian crossing, are clearly observable in perspective images. The ease of road learning from perspective images is well matched by the maturity of existing semantic segmentation models.
We adopt Mask2Former~\cite{mask2former} to first construct one-hot perspective road map $Map^p \in \mathbb{R}^{C{\times}H{\times}W}$, which assigns each pixel to the highest-probability semantic category based on class and mask probabilities. Here, $C$ represents the semantic category channel with a binary representation. Note that only the generated perspective road map is used.

\subsubsection{Inverse Perspective Mapping}
Since topological maps are constructed in BEV whereas the perspective road map is defined in a different view, a view transformer module is necessary. Inverse Perspective Mapping (IPM) is decoupled from model training, ensuring stable initial learning and generalization capability; it is therefore adopted as the view transformation strategy in this work. IPM assumes a flat road surface with zero ground elevation, allowing direct projection of perspective
information into the BEV. Given multi-view perspective road maps $Map_n^p$ with intrinsic $T_{n}^{in}$ and extrinsic $T_{n}^{ec}$ parameters, the BEV road structure map $Map_{bev}$ is obtained through a hypothetical ground-plane height $Z_0=0$:
\begin{equation}
    Map_{bev} = \sum\nolimits_{n} \text{IPM}(Map_n^p,T_{n}^{in},T_{n}^{ec},Z_0),
\end{equation}
where the summation fuses overlapping camera views additively, accumulating per-class evidence across cameras.

\input{figure/framework}
\subsection{Overview}
Topological mapping encompasses the detection of traffic sign and centerline instances, view transformer, and topological relationship learning. In light of its efficiency and low computational cost, TopoLogic~\cite{topologic} is chosen as the baseline architecture for this work.
The overview of the proposed HGeo-TopoMap is shown in Fig.~\ref{Fig:frame}. 

Given multi-view images $I_i$, multi-scale image features $F_i$ are initially extracted through an image encoding backbone. 
The detection of traffic sign instances is performed by a dedicated task head implemented with Deformable DETR~\cite{zhu2020deformable}. Concurrently, the image features are projected from the perspective view to BEV via BEVFormer~\cite{li2022bevformer}. 
The map instance decoder module extracts the centerline instance from the BEV features $F_{bev}$, which employs a DETR architecture to perform decoding from instance queries $Q_l$. Additionally, a graph convolutional network is integrated into the forward feedback network to capture the topological connections between centerline queries $Q_l$ and traffic queries $Q_t$.
Finally, the multi-task heads predict the topology matrix, centerline categories, and centerline points, respectively.

Given the lack of distinctive visual cues for constructing maps, this work incorporates multi-level geometric priors to boost topological mapping. 
A geometric adaptive learning module based on BEV road structure maps (introduced in Sec.~\ref{GAL}) and a geometric consistency learning module with spatial prior relationships (described in Sec.~\ref{GCL}).

\subsection{Geometric Adaptive Learning}
\label{GAL}
Road structure maps provide explicit geometric cues that are closely associated with centerline locations and configurations. For instance, centerlines are generally located between lane boundaries, while stop lines provide additional information about road intersections and driving directions. Such structural information can compensate for the absence of explicit centerline markings in perspective images. To obtain these priors, we first employ semantic segmentation to extract road structures from perspective images and subsequently project them into BEV using IPM. However, the resulting road structure maps may contain inaccuracies arising from segmentation ambiguity and occlusions~\cite{uncer}, as well as geometric distortions caused by the flat-ground assumption of IPM~\cite{genmapping}. Directly incorporating these imperfect priors may introduce irrelevant information into centerline representations. To address this issue, we develop a Geometric Adaptive Learning (GAL) module that adaptively aggregates spatially relevant road-structure information for each centerline instance.

\noindent\textbf{Multi-category Road Structure Encoding.}
Different road structure categories exhibit distinct semantic and spatial relationships with centerlines. To preserve these specific characteristics, we introduce a multi-category encoder that transforms the road structure map into semantic and positional representations. Specifically, given the IPM-derived road structure map $Map_{bev}$, we sample grid cells belonging to valid semantic categories to obtain their occupancy representations $M_o \in \mathbb{R}^{N_o \times C}$ and spatial coordinates $M_p \in \mathbb{R}^{N_o \times 2}$, where $N_o$ denotes the number of occupied grid cells and $C$ is the number of semantic categories:
\begin{equation}
M_o, M_p = \mathrm{Sample}(Map_{bev}, Map_{bev}>0).
\end{equation}
Here, $\mathrm{Sample}(\cdot,\cdot)$ extracts grid cells satisfying the specified occupancy condition. The sampled occupancy representations are then transformed into semantic embeddings through a multi-category encoder $\mathrm{MCE}(\cdot)$, while their spatial coordinates are encoded using a sinusoidal positional encoder $\mathrm{SinPos}(\cdot)$:
\begin{equation}
M_{emb} = \mathrm{MCE}(M_o), \quad
M_{pos} = \mathrm{SinPos}(M_p).
\end{equation}
The resulting road-structure tokens combine category-specific semantic information with spatial positional cues, providing explicit geometric priors for subsequent instance decoding.

\noindent\textbf{Distance-aware Prior Attention.}
Although road structure maps contain useful geometric information, not all structural regions are equally relevant to a given centerline instance. We therefore introduce a distance-aware attention mechanism that exploits the spatial proximity between predicted centerline points and occupied road-structure grids to guide instance-specific prior aggregation.

Let $P_I \in \mathbb{R}^{N_l \times N_p \times 3}$ denote the predicted centerline points, where $N_l$ and $N_p$ represent the number of centerline instances and the number of points per instance, respectively. We extract their planar coordinates $P_I' \in \mathbb{R}^{N_l \times N_p \times 2}$ and detach them from the computational graph when constructing the spatial guidance. For each centerline instance, we calculate the Euclidean distances between its predicted points and all occupied road-structure grids. The distance from an occupied grid to a centerline instance is defined as the minimum distance to any point belonging to that instance:
\begin{equation}
D_m(i,j) =
\min_{1 \leq t \leq N_p}
\left|
P_I'(i,t)-M_p(j)
\right|_2,
\end{equation}
where $D_m \in \mathbb{R}^{N_l \times N_o}$ represents the instance-to-structure distance matrix. This formulation captures the spatial relevance of road-structure regions with respect to the predicted geometry of each centerline instance.

Instead of applying a fixed spatial mask, we convert the distance matrix into a continuous attention bias:
\begin{equation}
B_{geo}=-kD_m,
\end{equation}
where $k$ is a scaling factor controlling the influence of spatial proximity. The resulting bias assigns relatively higher attention preference to road-structure tokens located closer to the predicted centerline geometry, while reducing the influence of spatially distant regions.

Given the centerline instance queries $Q_l$, the encoded road-structure features are used as keys and values in cross-attention. The distance-aware prior attention is formulated as
\begin{equation}
A_{\mathrm{geo}} =
\operatorname{Softmax}\left(
\frac{QK^{\top}}{\sqrt{d}} - kD_m
\right),
\label{eq:gal_attn}
\end{equation}
where $Q=Q_lW_Q$, $K=(M_{emb}+M_{pos})W_K$, and $d$ denotes the feature dimension. The geometry-guided query features are then obtained as $Q_{\mathrm{geo}}=A_{\mathrm{geo}}V$, where $V=M_{emb}W_V$.

By jointly considering query-feature similarity and geometric proximity, GAL enables each centerline query to selectively aggregate spatially relevant road-structure information. The resulting geometry-guided query representations are subsequently refined through query self-attention and cross-attention with BEV visual features, providing structural guidance for centerline instance decoding.

\subsection{Geometric Consistency Learning}
\label{GCL}

While GAL incorporates explicit geometric guidance from road structure maps, centerline instances themselves exhibit intrinsic geometric regularities that can further facilitate representation learning. As illustrated in Fig.~\ref{fig:intro}, centerlines can be broadly categorized into straight and curved instances according to their geometric configurations. Moreover, straight centerlines often exhibit characteristic orientation patterns, such as approximately parallel arrangements in adjacent lanes. These geometric properties provide complementary guidance for distinguishing centerline instances. However, conventional point regression does not explicitly model the relative geometric configuration of centerline points, while standard detection objectives provide limited supervision for learning geometric similarities among instances. To address these limitations, we introduce a Geometric Consistency Learning (GCL) module that combines center-guided hierarchical decoding with geometry-aware contrastive supervision, facilitating the learning of structured and geometrically consistent instance representations.

\input{figure/pointreg}

\noindent\textbf{Center-guided Hierarchical Decoding.}
To facilitate geometric representation learning, we first introduce a center-guided point regression strategy that explicitly models the relative spatial configurations of centerline instances. Unlike direct point regression from reference points~\cite{topologic}, the proposed strategy decomposes centerline prediction into instance center localization and center-relative point regression within each decoder layer.

Specifically, given the instance queries $Q_l$ and their corresponding reference points $P_r$, an instance center regression head first predicts the center locations:
\begin{equation}
    P_c = \mathrm{CenRegHead}(Q_l) + P_r,
\end{equation}
where $\mathrm{CenRegHead}(\cdot)$ consists of linear, normalization, and activation layers. Subsequently, an instance point regression head predicts the relative offsets of individual centerline points with respect to their corresponding instance centers:
\begin{equation}
    P_I = \mathrm{InsRegHead}(Q_l) + P_c,
\end{equation}
where $\mathrm{InsRegHead}(\cdot)$ has a structure similar to that of the center regression head, and $P_c$ is broadcast to all predicted points within each instance.

By introducing instance center information into both point regression and feature interaction, the proposed decoding strategy facilitates the learning of centerline geometry within each decoder layer. As illustrated in Fig.~\ref{Fig:pointreg}(b), centerline instances with similar orientations exhibit consistent relative arrangements of their predicted points, providing a structured representation for subsequent geometric consistency learning.

\noindent\textbf{Geometry-aware Contrastive Learning.}
Building upon the center-guided decoding strategy, we further introduce geometry-aware contrastive supervision to encourage instance queries with similar geometric attributes to learn consistent representations. 

We first categorize ground-truth centerlines into straight and curved groups, denoted by $G_l$ and $G_c$, respectively, using a three-point discrete curvature estimator with threshold $t_c$. Since straight centerlines may exhibit different spatial orientations, we further partition them into orientation-aware subgroups to capture their geometric relationships. 
Specifically, the slopes of straight centerlines are computed in the vehicle coordinate system and sorted in ascending order. The gaps between adjacent slopes are then used to identify orientation boundaries, allowing centerlines with similar orientations to be grouped together. The resulting geometric groups are defined as
\begin{equation}
G=[G_{l1},\ldots,G_{ln},G_c],
\end{equation}
where $n$ denotes the number of orientation-based subgroups for straight centerlines.
The influence of different grouping granularities is investigated in the ablation study.

This geometric attribute grouping operates on the ground truth, where the attributes of each predicted instance are derived from its matched ground-truth counterpart.
To guarantee that instance features of each geometric property are fully utilized for guidance, the contrastive learning scheme is concurrently applied to all groups.
The anchor plays a central role in guiding the learning direction in contrastive learning. 
Thus, instance queries that are correctly matched to the ground truth are designated as the anchor set within each subgroup $G_v$:  
\vspace{-0.5em}
\begin{equation}
    v = \{Q_l|Q_l \in G_v,\text{match}(Q_l,\text{GT})=\text{True}\}.
\end{equation}
Positive samples $S_{pos}$ are selected from other high-scoring instances within the same group.
Negative samples $S_{neg}$ are drawn from other groups $G_{other}$ with distinct geometric features.  Low-scoring instances are discarded in both cases to avoid misleading guidance.
\begin{equation}
    S_{pos} = \{Q_l|Q_l\in G_v,e_l>t_s\},
\end{equation}
\begin{equation}
    S_{neg} = \{Q_l|Q_l\in G_{other},e_l>t_s\},
\end{equation}
The instance reliability threshold $t_s$ follows the same criteria used in other works~\cite{toponet,topologic} for selecting reliable instances.
$e$ is the score of each instance.

A lightweight projection head consisting of linear layers maps the selected instance queries into a contrastive embedding space. Let $z_a$, $z_p$, and $z_n$ denote the projected embeddings of an anchor, a positive sample, and a negative sample, respectively. We adopt a multi-positive contrastive objective:
\begin{equation}
\mathcal{L}_{gcl}
=
-\sum_{a\in\mathcal{A}}
\log
\frac{
\sum_{p\in\mathcal{S}_{pos}(a)}
\exp(z_a^\top z_p)
}{
\Omega(a)
},
\end{equation}
where
\begin{equation}
\begin{aligned}
\Omega(a)={}&
\sum_{p\in\mathcal{S}_{pos}(a)}
\exp(z_a^\top z_p)\\
&+\sum_{n\in\mathcal{S}_{neg}(a)}
\exp(z_a^\top z_n).
\end{aligned}
\end{equation}
Here, $\mathcal{A}$ denotes the set of valid anchor queries.
By combining instance-centered geometric decoding with layer-wise contrastive supervision, GCL facilitates the learning of geometric orientations and consistent instance representations, complementing the explicit road-structure guidance provided by GAL.

\subsection{Overall Loss}
The standard topological mapping task involves three loss terms, namely traffic detection loss $L_{det}$, centerline instance detection loss $L_{l}$, and topology reasoning loss $L_{topo}$, which are adopted similarly to TopoLogic and other existing methods.
Based on this foundation, an instance center point loss $L_c= \text{L1}(P_c,GT_c)$.
where $GT_c$ is the center point of each centerline instance GT.
The complete loss function is:
\begin{equation}
    L = L_{det} + L_{l} + L_{topo}+\alpha *L_c+\beta *L_{gcl},
\end{equation}
where $\alpha$ and $\beta$ denote the loss weights for center-point regression and contrastive learning, respectively.

%% file: figure/framework.tex
\begin{figure*}[t]
      \centering
      \includegraphics[scale=0.5]{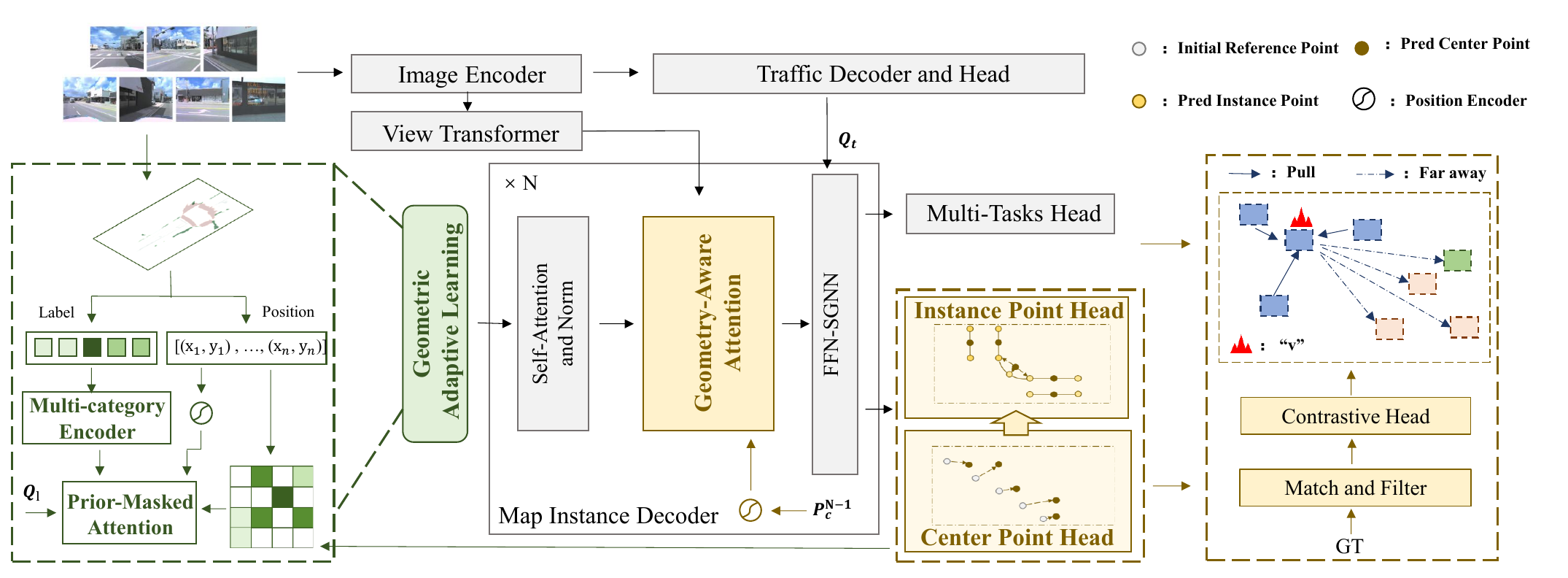}
      \caption{Overview of the proposed HGeo-TopoMap architecture.
      The proposed geometric adaptive learning module embeds road structure map priors into instance decoding, while the geometric consistency learning module uses geometric contrastive learning for implicit geometric relationships.
      }
      \label{Fig:frame}
\end{figure*}

%% file: figure/pointreg.tex
\begin{figure}[!t]
      \centering
      \includegraphics[scale=0.42]{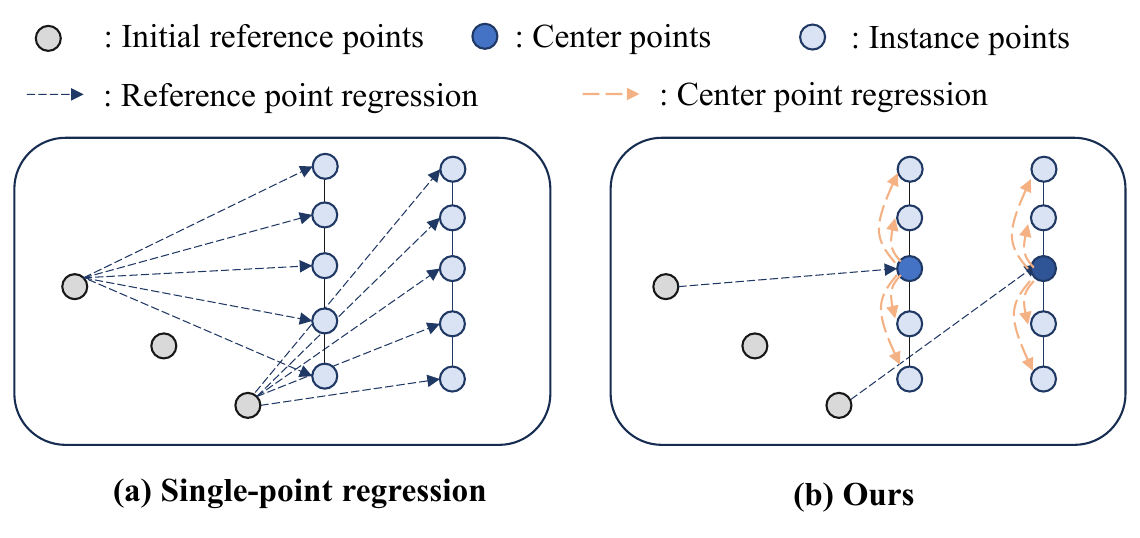}
      \caption{A comparison of instance point regression learning across various decoders. 
      }
      \label{Fig:pointreg}  
\end{figure}

%% file: contents/experiments.tex
\input{figure/roadmap}
\input{table/t1-2}
\input{figure/show_a}
\subsection{Datasets and Metrics}
OpenLane-V2~\cite{wang2023openlanev2} is a large-scale dataset established for topological reasoning in autonomous driving.
The dataset offers diverse annotations, covering centerlines, traffic signs, inter-centerline topology, and centerline-traffic sign topology.
Additionally, LaneSegNet~\cite{lanesegnet} provides auxiliary lane annotations for each centerline.
The proposed method is evaluated on the centerline and lane segment benchmark of this dataset.

OpenLane-V2 provides a standardized evaluation protocol. The accuracy of centerline instances is measured by $\text{DET}_l$, which is the average precision across different Fréchet distance thresholds.
Traffic sign detection $\text{DET}_t$ adopts Intersection over Union (IoU) as the similarity metric. $\text{TOP}_{ll}$ and $\text{TOP}_{lt}$ are used to evaluate the similarity of topology matrices among lanes and between lanes and traffic elements, respectively. Finally, the v2.1.0 of $\text{OLS}$ is employed to measure overall topological map accuracy.

In the lane segment benchmark, it is evaluated using the same metrics as LaneSegNet. 
Average precision is adopted for both lane $\text{AP}_{ls}$ and pedestrian $\text{AP}_{ped}$ detection, while $\text{TOP}_{lsls}$ is used for topological reasoning.

\subsection{Implementation Details}
The input size for multi-angle images is $800\times1024$, which is encoded by the ResNet50 model.
The range for topological mapping is $102.4\text{m}{\times}51.2$m, corresponding to BEV features of size $200{\times}100$.
And centerline instances are represented as polylines.
The perspective road maps with a $388{\times}512$ size are generated by Mask2Former, which is pre-trained on the Mapillary Vistas dataset~\cite{Mapillary}.
During training, the road structure map is generated in real time. Its range matches the topological map, but sampling is limited to $60\text{m}{\times}30$m, with remaining regions set to zero. 
This is because IPM suffers from considerable noise at long range due to occlusions and other factors, as shown in Fig.~\ref{Fig:road}. Compared with the relatively accurate mapping at close range, the distant regions contain substantial noise.
The fixed parameter is defined as $k{=}0.1$, $t_c{=}0.05$ and $t_s{=}0.3$.
The weights of the loss are set as $\alpha{=}1$ and $\beta{=}0.1$. The former uses the same weight as the point regression loss; the latter is determined by ablation.
All experiments are trained for $24$ epochs on NVIDIA RTX A6000 GPUs with a batch size of $4$.
The optimizers and learning rate for different tasks remain consistent with the respective baselines.

\input{table/t2}
\input{table/t3}
\subsection{ Comparison on OpenLane-V2 Dataset}
\textbf{Comparison on the Centerline Benchmark:}
This evaluates centerline and traffic element detection, together with two topology reasoning tasks. Table~\ref{tab:center_compare} presents the results on subset A, including representative baseline methods and recent topological mapping approaches. HGeo-TopoMap is designed as a flexible geometric enhancement framework that can be integrated into different topological mapping architectures. To evaluate its generalizability, we incorporate the proposed modules into three representative baselines, namely TopoLogic~\cite{topologic} and TopoPoint~\cite{fu2026topopoint}. The results show that HGeo-TopoMap consistently improves topological mapping performance across these architectures, demonstrating its effectiveness and architectural flexibility. In particular, integrating HGeo-TopoMap into TopoLogic yields a $+2.0$ improvement in the OLS metric.

To further investigate the effectiveness of geometric prior guidance under different scene complexities, we partition the original validation set into two splits according to the number of centerlines. The Hard subset contains scenes with three or more centerlines, while the remaining scenes constitute the Easy subset. As shown in Table~\ref{tab:center_compare}, the proposed method achieves more pronounced improvements on the Hard subset, suggesting that geometric priors are particularly beneficial for centerline detection in complex multi-lane scenarios.

Furthermore, Table~\ref{tab:center_compare_b} reports the results on subset B, where HGeo-TopoMap achieves a $+1.9$ improvement in OLS over the corresponding baseline. These results further support the effectiveness and transferability of the proposed geometric enhancement framework across different evaluation subsets.

\textbf{Comparison on the Lane Segment Benchmark:}
This benchmark includes auxiliary tasks for lane and pedestrian detection, as originally proposed by LaneSegNet~\cite{lanesegnet}.
The comparison results under identical configurations and environments are shown in Table~\ref{tab:lane_compare}. 
LaneSegNet+Ours framework achieves an instance accuracy of $34.0$ mAP. This corresponds to an improvement of $+2.3$ mAP over the baseline.
These results suggest that hierarchical geometric prior guidance can benefit both centerline instance detection and topological mapping.
Moreover, the inter-lane topology reasoning accuracy improves by $+1.8$.

\input{table/core_module}
\input{table/position}
\subsection{Ablation Studies}
\textbf{Core Modules:}
This section evaluates the effectiveness of the two proposed modules. The results are shown in Table~\ref{tab:core}. 
First, the GCL and GAL modules individually improve the OLS metric by $+0.8$ and $+1.3$, respectively.
When both are applied together, the OLS metric improves by $+2.0$.
While the improvements are not strictly monotonic across every individual metric (\textit{e.g.}, $\text{Top}_{lt}$ decreases by 0.1), the OLS of $45.5$ supports the complementary effect of the two components.

\textbf{Position of GAL Module:}
The BEV road structure map is integrated into the iterative centerline instance decoding pipeline. 
We ablate its insertion position, where decoder layers are indexed from $0$.
Table~\ref{tab:position} reports results on two baselines, which show different trends. 
The lane segment baseline reports that embedding the map at the intermediate decoder layer yields optimal results, achieving a centerline detection accuracy of $33.7$ mAP. 
But the centerline baseline performs best with early embedding. 
This discrepancy can be attributed to their different decoding paradigms.
The former uses regression-based learning, where early features are unaligned. 
The latter directly predicts instances at each layer, making early injection more effective for spatial correction.

\input{table/dismask}
\input{figure/show_b}
\input{table/weight_loss}
\textbf{Design of GAL Module:}
This section further examines the distance mask of the GAL module.
Table~\ref{tab:dismask} shows that merely adding the mask to cross-attention yields no performance gain, mainly because learning distance masks from predicted reference points is uncertain. To regulate mask-guided learning, we scale the mask distance by $k$. 
When $k$ is $0.1$, the accuracy reaches its optimum, where mAP is $28.5$ and topological reasoning is $19.5$.

\textbf{Loss Weight of GCL Module:}
The GCL module adopts the consistent loss with a weight factor $\beta$ to constrain the learning of spatial relationships.
Table~\ref{tab:lossweight} demonstrates that the optimal overall topological map accuracy is achieved with a weight value of $0.01$. 
However, when the weight is set to $0.1$, traffic sign accuracy slightly decreases, while instance detection accuracy remains comparable.
In this case, topological reasoning achieves a more balanced performance.  

\input{table/sample}
\textbf{Sample Selection of GCL Module:}
Given the spatial distribution differences in multi-intersection scenes, this section validates the learning effectiveness under different maximum groupings.
Table~\ref{tab:sample} shows that finer groupings do not yield further accuracy gains, indicating that two orthogonal directions suffice globally.
Finer groupings instead introduce sampling noise, whereas the two-group strategy achieves the best trade-off between geometric discriminability and sampling robustness.

\input{table/center}
\input{table/param}
\textbf{Structure of GCL Module:}
The GCL module consists of a center-guided hierarchical decoding and a geometric-aware contrastive learning component. In this section, we evaluate their individual effectiveness separately.
Table~\ref{tab:center} reports that both components are effective.
Compared with geometric-aware contrastive learning alone, joint learning with the center-guided hierarchical decoding further unleashes its potential, yielding a $0.8$ gain in the OLS metric.

\input{figure/show_c}
\textbf{Discussion on Model Efficiency and Parameter Amount:}
Table~\ref{tab:param} compares the model complexity and efficiency of different methods. The proposed method increases the trainable parameter count by only $2.4$M, indicating that the performance gain does not primarily stem from increased trainable capacity, but rather from the hierarchical geometric priors.
During inference, the road structure map consumed by GAL is generated by a semantic segmentation model, and the inference speed of the proposed framework is therefore governed by that of the segmentation model. To characterize this dependency, we evaluate two segmentation backbones, ResNet-50 and Swin-L, under identical mapping configurations.
The full pipeline is slower than the baseline but consistently more accurate. With the lighter R50, the speed rises to $5.0$ FPS while the OLS changes only marginally, showing that the segmentation front-end offers a tunable accuracy–efficiency trade-off for our framework.
In future work, we plan to explore using road structure maps from historical frames within a sliding window as priors to improve real-time efficiency. In this case, the pipeline speed can be substantially improved.

\subsection{Visual Results}
The visualization results of the centerline benchmark on OpenLane-V2 subsets A and B are shown in Fig.~\ref{Fig:show_a}. 
The experimental results demonstrate that the proposed method achieves a more detailed mapping capability.
Notably, in complex scenarios such as multi-lane merging, the proposed method identifies centerline structures more accurately. This is consistent with the results in Table~\ref{tab:center_compare}, where the proposed method yields more pronounced gains on the multi-lane hard-split data. This capability is crucial for the safe deployment of autonomous vehicles, as merging areas are typically associated with higher collision risks.

Meanwhile, the visualization result of the lane segment benchmark is shown in Fig.~\ref{Fig:show_b}.
Obviously, compared to LaneSegNet~\cite{lanesegnet}, the proposed method achieves a more precise reconstruction of road elements.
This is attributed to the semantic prior in the road structure map, which effectively guides instance learning.
Furthermore, the proposed method is more rigorous in identifying fine details of road structures.

\input{figure/robust}
\subsection{Robustness Analysis}
Multi-view camera topological maps depend on view completeness for robustness, whereas our IPM-based road structure map relies on the road planarity assumption.
We accordingly analyze robustness from these two dimensions.

\textbf{Robustness under Sensor Damage:}
Sensor damage in real scenes often causes missing views, endangering vehicle safety.
To comprehensively assess the robustness of the proposed method, 
this section evaluates map construction performance under missing viewpoints.
Note that the corresponding regions of the 
road structure map are also set to contain no information.
As shown in Fig.~\ref{Fig:robust}, experiments under various missing-view scenarios reveal that the absence of front-view cameras causes the most significant performance degradation. This finding is consistent with the inherent reliance of autonomous driving on forward perception.
Nevertheless, the proposed method demonstrates significantly better robustness than the baseline. 
Notably, under the rear-left view missing setting, the proposed method achieves a $7.5$ improvement over the baseline on $\text{DET}_l$.

Moreover, the visualization comparison is presented in Fig.~\ref{Fig:show_c}. Owing to the missing views, the instances learned by the map instance decoder are correspondingly noisy, which prevents accurate instance detection. Benefiting from the guidance of prior information, the proposed method exhibits enhanced robustness under missing views relative to TopoLogic and is capable of detecting complete road structures more effectively.
Both normal and missing-view experimental results consistently demonstrate that the proposed method is more applicable to real-world scenarios.

\textbf{Sensitivity of IPM to Environmental Changes:}
Since Inverse Perspective Mapping (IPM) is based on the flat-road assumption, this section analyzes the sensitivity of IPM under different environments.
Table~\ref{tab:noise} shows that the road structure map also performs well at long distances. Because road structures are extensible, a local prior can effectively guide long-range learning.
Moreover, the averaged results under ground heights or parameter perturbations further demonstrate that the proposed method maintains stable map construction, verifying the robustness of the geometric adaptive learning.

\input{table/noise_ex}

\input{figure/night_show}

\subsection{Challenging Scenarios}
Real-world images under different weather conditions exhibit notable domain gaps, making it challenging for models trained on mixed domains to learn the feature distributions of extreme nighttime scenes. 
As shown in Fig.~\ref{fig:night}, in the nighttime scenes of subset B, TopoLogic~\cite{topologic} suffers from considerable missed detections of centerline instances. The proposed method leverages multi-level priors to improve generalization and thus achieves relatively better detection capability; however, it still exhibits errors in identifying distant centerlines.
This further suggests that domain adaptation research is worth exploring to improve real-world deployment generalization in complex unconstrained surroundings under adverse conditions.

%% file: figure/roadmap.tex
\begin{figure}[!t]
      \centering
      \includegraphics[scale=0.36]{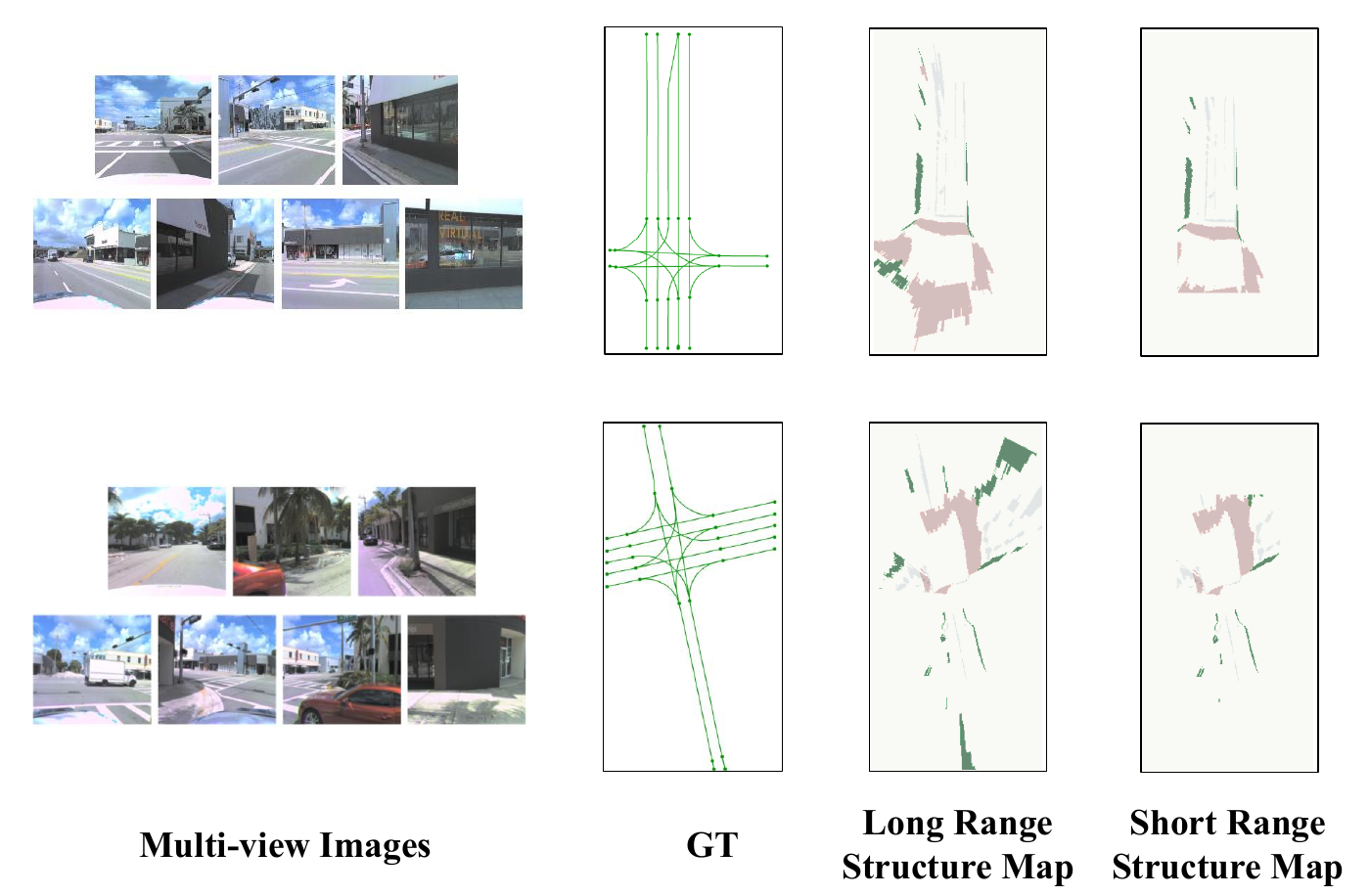}
      \caption{Visualization of road structure maps. The road structure map generated via IPM is more reliable at close range and can be aligned with the topological map, whereas the map structure at long range contains considerable noise.
      }
      \label{Fig:road}  
\end{figure}

%% file: table/t1-2.tex
\begin{table*}[t]
\fontsize{8.5}{12}\selectfont
\renewcommand{\arraystretch}{1.1}
\setlength\tabcolsep{9pt}
\caption{Results of the centerline benchmark on the OpenLane-V2 subset A dataset~\cite{wang2023openlanev2}. 
$*$ denotes reproduced results. 
}
\label{tab:center_compare}
\begin{center}

\begin{tabular}{l|c|ccccc|c|c}
\hline
 \multirow{2}{*}{Method }  &\multirow{2}{*}{Venues }    &\multicolumn{5}{c|}{Ori-Split}  &\multicolumn{1}{c|}{Simple-Split}&\multicolumn{1}{c}{Hard-Split} \\ \cline{3-9}
 & & $\text{DET}_{\textit{l}}$ & $\text{DET}_{\textit{t}}$ & $\text{TOP}_{\textit{ll}}$ & $\text{TOP}_{\textit{lt}}$ & OLS  & OLS & OLS \\ \hline \hline
TopoNet~\cite{toponet} &Arxiv~2023  & 28.6   & 48.6   & 10.9    & 23.8    & 39.8 &--&--\\
TopoMLP~\cite{topomlp} &ICLR~2024 & 28.8   & 53.3   & 7.8     & 30.1    & 41.2 &-- &--\\
TopoLogic*~\cite{topologic} &NeurIPS~2024  & 29.7        & 43.6       & 25.0    & 25.5     & 43.5 &41.9 &43.3  \\ 
Topo2D~\cite{topo2d}&Arxiv~2024 & 29.1   & 50.6   & 22.3    & 26.2    & 44.5 &-- &--  \\ 
TopoPoint*~\cite{fu2026topopoint}&NeurIPS~2025 &{29.8} &{54.3} &{28.5} &{29.7} &{48.0} &47.1 &48.3\\
UniTopo*~\cite{unitopo} &TCSVT~2026 &30.4  &48.9   &29.5  &31.2  &47.4 &47.3 & 47.3 \\
\hline
TopoLogic+Ours &--  &  31.3 & 48.2   & 26.2  & 26.3  & 45.5 &43.1 & 45.4  \\ 
TopoPoint+Ours &-- &31.0 &55.9 &28.9 &29.9 &48.8 &47.2 &49.0 \\
\hline
\end{tabular}
\end{center}
\end{table*}

%% file: figure/show_a.tex
\begin{figure*}[!t]
      \centering
      \includegraphics[scale=0.34]{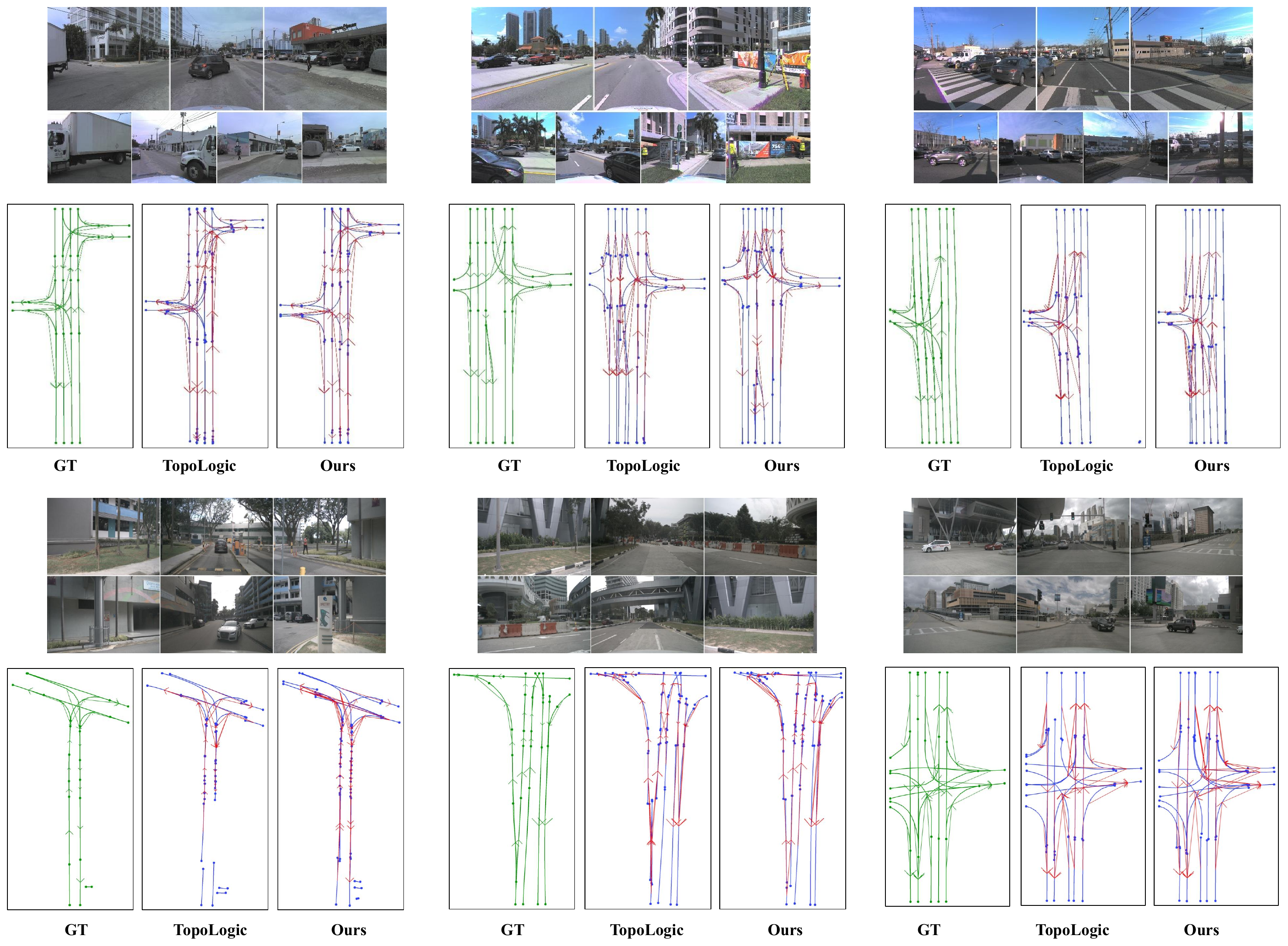}
      \caption{Qualitative results of the centerline benchmark across different scenarios on OpenLane-V2 subsets A and B~\cite{wang2023openlanev2}. 
      Compared with TopoLogic~\cite{topologic}, the proposed method identifies centerlines more accurately.
      }
      \label{Fig:show_a}  
\end{figure*}

%% file: table/t2.tex
\begin{table}[t]
\fontsize{8}{10}\selectfont
\renewcommand{\arraystretch}{1.1}
\setlength\tabcolsep{9pt}
\caption{Results of the centerline benchmark on OpenLane-V2 subset B~\cite{wang2023openlanev2}.
$*$ denotes reproduced results. }
\vspace{-1.em}
\label{tab:center_compare_b}
\begin{center}
\resizebox{1.0\linewidth}{!}{\begin{tabular}{l|cccccc}
\hline
Method      & $\text{DET}_{\textit{l}}$ & $\text{DET}_{\textit{t}}$ & $\text{TOP}_{\textit{ll}}$ & $\text{TOP}_{\textit{lt}}$ & OLS \\ \hline \hline
 TopoNet~\cite{toponet} &24.3 &55.0 &6.7 &16.7 &36.8 \\
 TopoMLP~\cite{topomlp} &26.6 &58.3 &7.6 &17.8 &38.7 \\
TopoLogic*~\cite{topologic} &{24.5} &{56.0} &{19.8} &{16.4} &{41.4}    \\ 
\hline
TopoLogic+Ours    & {28.8}      &{53.0}    &{24.2}    & {18.0}     &{43.3}    \\ 
\hline
\end{tabular}}
\end{center}
\end{table}

%% file: table/t3.tex
\begin{table}[t]
\fontsize{7}{9}\selectfont
\renewcommand{\arraystretch}{1.1}
\setlength\tabcolsep{9pt}
\caption{Results of the lane segment benchmark on the OpenLane-V2 subset A~\cite{wang2023openlanev2}. 
\textsuperscript{+} represents the official weight verification result.}
\label{tab:lane_compare}
\begin{center}
\resizebox{1.0\linewidth}{!}{\begin{tabular}{l|cccc}
\hline
Method   & mAP  & $\text{AP}_{ls}$ & $\text{AP}_{ped}$ & $\text{TOP}_{lsls}$ \\ \hline \hline
TopoNet~\cite{toponet}              & 23.0 & 23.9    & 22.0     & 1.0       \\
MapTR~\cite{liao2022maptr}                  & 27.0 & 25.9    & 28.1     & --        \\
MapTRv2~\cite{maptrv2}             & 28.5 & 26.6    & 30.4     & --        \\
LaneSegNet\textsuperscript{+}~\cite{lanesegnet}    & 31.7 & 28.4    & 34.9     & 20.8      \\\hline
LaneSegNet+Ours   & {34.0}   & {32.3}    & {35.7}  & {22.6}           \\ \hline
\end{tabular}}
\end{center}
\end{table}

%% file: table/core_module.tex
\begin{table}[t]
\fontsize{7}{8}\selectfont
\renewcommand{\arraystretch}{1.1}
\setlength\tabcolsep{9pt}
\caption{Ablation results of the core modules.}
\vspace{-1.0em}
\label{tab:core}
\begin{center}
\resizebox{1.0\linewidth}{!}{\begin{tabular}{cc|ccccc}
\hline
GCL & GAL & $\text{DET}_l$ & $\text{DET}_t$ & $\text{TOP}_{ll}$ & $\text{TOP}_{lt}$  & OLS       \\
\hline \hline
    &     & 29.7        & 43.6       & 25.0    & 25.5     & 43.5      \\
 \checkmark   &     & 30.6        & 44.4       & 26.1    & 26.2     & 44.3      \\
   &  \checkmark     & 31.2        &  45.8       & 25.9     & 26.4      &  44.8      \\
  \checkmark   & \checkmark     & 31.3        & 48.2       & 26.2    & 26.3     & 45.5      \\ \hline     
\end{tabular}}
\end{center}
\end{table}

%% file: table/position.tex
\begin{table}[!t]
\fontsize{8}{9}\selectfont
\renewcommand{\arraystretch}{1.1}
\setlength\tabcolsep{9pt}
\caption{Ablation results of the fusion position.}
\label{tab:position}
\begin{center}
\resizebox{1.0\linewidth}{!}{\begin{tabular}{lc|ccccc}
\hline
GAL &Layer & \multicolumn{2}{c}{mAP}  & $\text{AP}_{ls}$ & $\text{AP}_{ped}$ & $\text{TOP}_{lsls}$ \\ \hline \hline
\checkmark &1  & \multicolumn{2}{c}{31.7} & 29.9    & 33.5     & 22.2      \\
\checkmark &3  & \multicolumn{2}{c}{33.7} & 31.7    & 35.8     & 22.4      \\
\checkmark &5  & \multicolumn{2}{c}{32.4}    & 31.4        & 33.4         & 22.4          \\ \hline 
GAL &Layer & $\text{DET}_l$ & $\text{DET}_t$ & $\text{TOP}_{lt}$ & $\text{TOP}_{ll}$  & OLS \\ \hline \hline
\checkmark &1  & 31.3      & 48.2     & 26.2    & 26.3     & 45.5 \\     
\checkmark &3   &31.3 &47.7  &26.3   &25.7   &45.2 \\ \hline
\end{tabular}}
\end{center}

\end{table}

%% file: table/dismask.tex
\begin{table}[t]
\fontsize{7}{8}\selectfont
\renewcommand{\arraystretch}{1.1}
\setlength\tabcolsep{9pt}
\caption{Ablation results of different geometric adaptive learning methods under $12$ training epochs.}
\vspace{-1.0em}
\label{tab:dismask}
\begin{center}
\resizebox{1.0\linewidth}{!}{\begin{tabular}{lcc|cccc}
\hline

Method &Mask  &$k$ & mAP  & $\text{AP}_{ls}$ & $\text{AP}_{ped}$ & $\text{TOP}_{lsls}$ \\ \hline \hline
CrossAttn &--    &--   &26.8     & 25.2  & 28.3    &19.0 \\
CrossAttn &\checkmark    &1   &26.7     &24.2   &29.1     &18.6 \\
CrossAttn &\checkmark    &0.1   &28.5  &26.2     &30.9   &19.5  \\
CrossAttn &\checkmark    &0.01   &27.8  & 25.5    &30.1   &19.4  \\ 
\hline 
\end{tabular}}
\end{center}
\vspace{-1.5em}
\end{table}

%% file: figure/show_b.tex
\begin{figure*}[!t]
      \centering
      \includegraphics[scale=0.34]{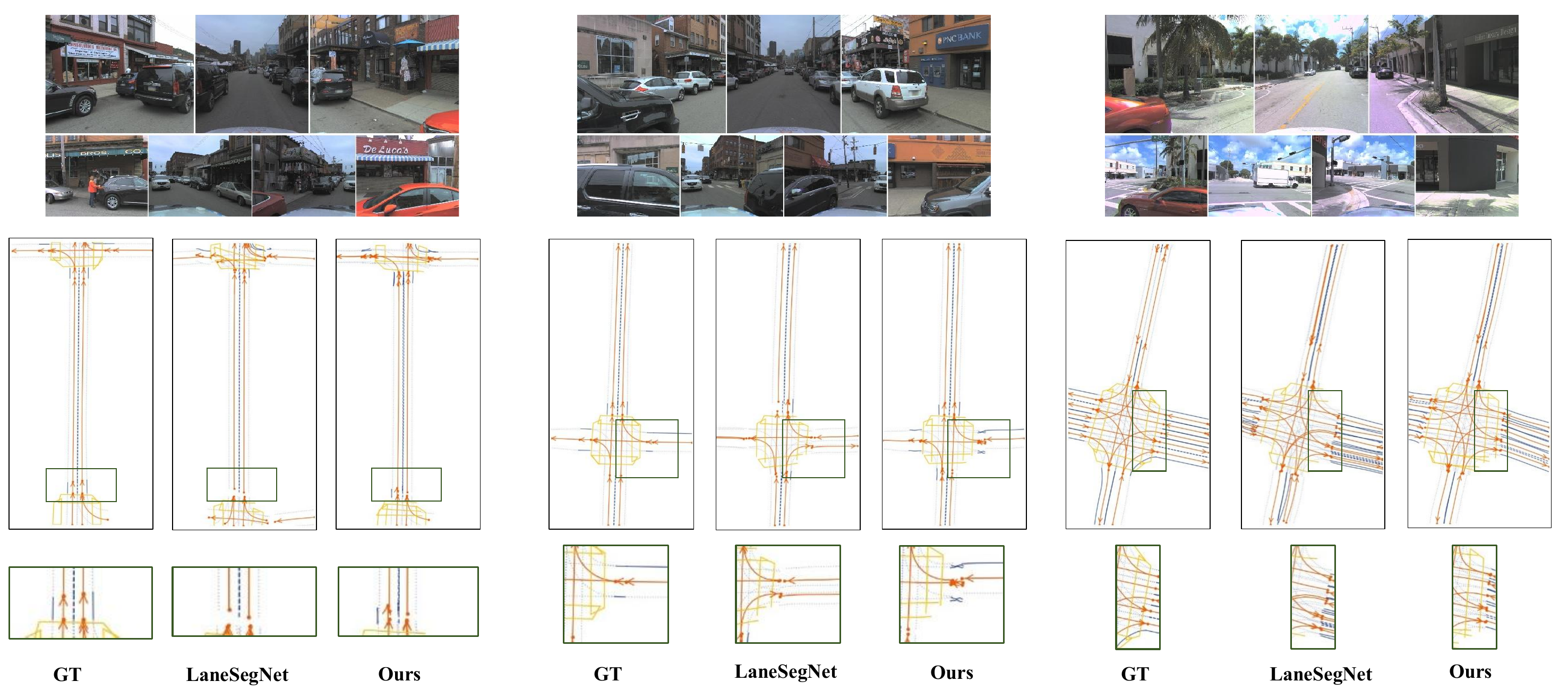}
      \caption{Qualitative results of the lane segment benchmark across different scenarios on OpenLane-V2 subset A. 
      }
      \label{Fig:show_b}  
\end{figure*}

%% file: table/weight_loss.tex
\begin{table}[t]
\fontsize{6}{8}\selectfont
\renewcommand{\arraystretch}{1.1}
\setlength\tabcolsep{9pt}
\caption{Ablation results of the loss weight for the GCL module.}
\label{tab:lossweight}
\begin{center}
\resizebox{1.0\linewidth}{!}{
\begin{tabular}{l|ccccc}
\hline
Weight & $\text{DET}_l$ & $\text{DET}_t$ & $\text{TOP}_{\textit{ll}}$ & $\text{TOP}_{\textit{lt}}$ & OLS  \\ \hline \hline
0.01   & 30.8   & 45.6   & 26.5    & 25.6    & 44.6 \\
0.1    & 30.6   & 44.4   & 26.1    & 26.2    & 44.3 \\ 
0.5   &30.0  &45.5    &23.4    &24.5   &43.3\\ \hline 
\end{tabular}}
\end{center}
\end{table}

%% file: table/sample.tex
\begin{table}[t]
\fontsize{7}{9}\selectfont
\renewcommand{\arraystretch}{1.1}
\setlength\tabcolsep{9pt}
\caption{Ablation results of the linear grouping.}
\label{tab:sample}
\begin{center}
\resizebox{1.0\linewidth}{!}{\begin{tabular}{lc|ccccc}
\hline
Cur. &Stra.  & $\text{DET}_l$ & $\text{DET}_t$ & $\text{TOP}_{ll}$ & $\text{TOP}_{lt}$ & OLS  \\ \hline \hline
1 &4     & 29.2     & 47.6     & 26.3  & 26.2       &  44.8   \\
1 &2      &  31.3 & 48.2   & 26.2  & 26.3  & 45.5       \\ \hline
\end{tabular}}
\end{center}
\vspace{-1em}
\end{table}

%% file: table/center.tex
\begin{table}[t]
\fontsize{7}{8.5}\selectfont
\renewcommand{\arraystretch}{1.1}
\setlength\tabcolsep{9pt}
\caption{Ablation results of the Center-guided Hierarchical Decoding (CGHD) and Contrastive Learning (CL) in the GCL module.}
\label{tab:center}
\begin{center}
\resizebox{1.0\linewidth}{!}{\begin{tabular}{lc|ccccc}
\hline
CGHD &CL & $\text{DET}_l$ & $\text{DET}_t$ & $\text{TOP}_{ll}$ & $\text{TOP}_{lt}$ & OLS  \\ \hline \hline
& & 29.7        & 43.6       & 25.0    & 25.5     & 43.5      \\
\checkmark & & 29.8        & 43.4       & 25.9    & 25.7     & 43.7    \\
&  \checkmark     & 30.0        & 44.7       &  25.3   & 25.3     & 43.8     \\
 \checkmark   &\checkmark     & 30.6        & 44.4       & 26.1    & 26.2     & 44.3         \\ \hline
\end{tabular}}
\end{center}
\end{table}

%% file: table/param.tex
\begin{table}[t]
\fontsize{10}{12}\selectfont
\renewcommand{\arraystretch}{1.1}
\setlength\tabcolsep{9pt}
\caption{Comparison of the number of parameters and efficiency.}
\label{tab:param}
\begin{center}
\resizebox{1.0\linewidth}{!}{\begin{tabular}{l|cccc}
\hline
Method & Seg Model& \# Training Params & FPS  &OLS  \\ \hline \hline
TopoLogic~\cite{topologic} &-- & 62.1M &6.4 &43.5 \\
TopoLogic+Ours &Mask2Former+R50& 64.4M  &5.0&45.4 \\
TopoLogic+Ours &Mask2Former+Swim-L & 64.4M & 3.6 &45.5  \\ 
\hline
\end{tabular}}
\end{center}
\end{table}

%% file: figure/show_c.tex
\begin{figure*}[!t]
      \centering
      \includegraphics[scale=0.34]{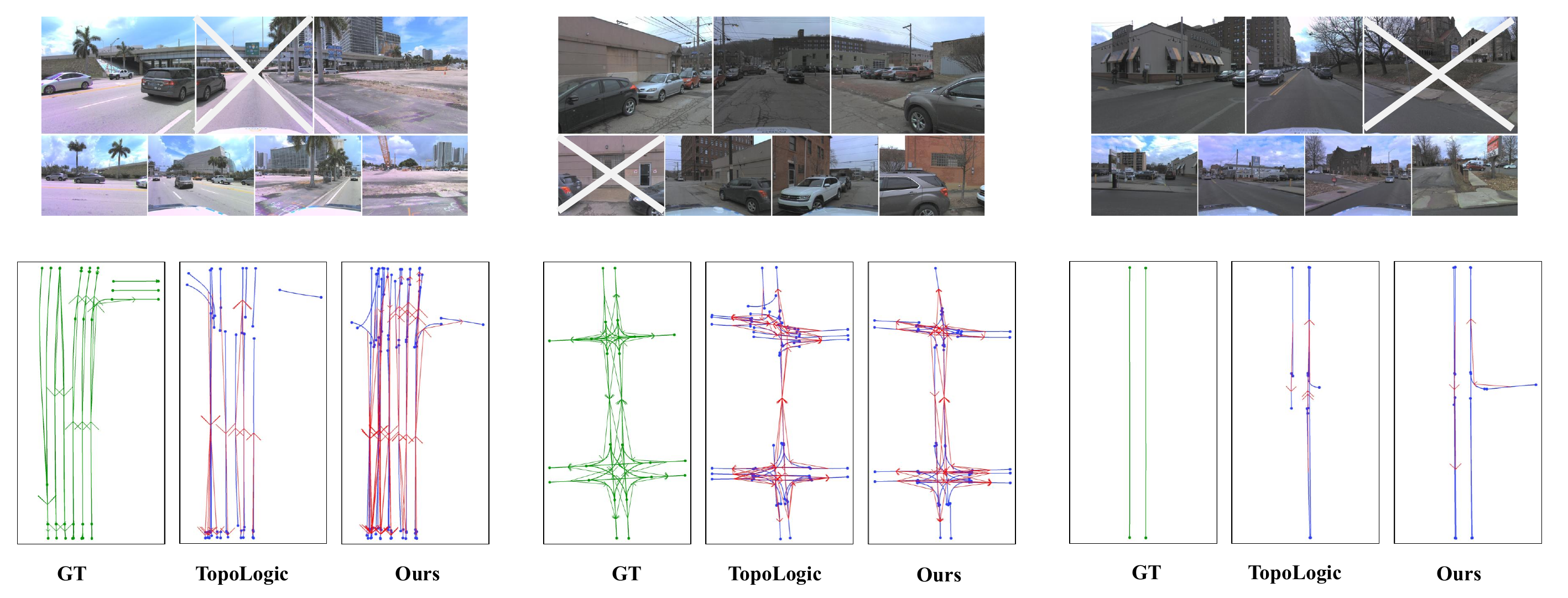}
      \caption{Visualization results under missing views. The figure compares TopoLogic~\cite{topologic} and the proposed method under the front view, the front-right view and the rear-left missing settings.
      }
      \label{Fig:show_c}  
\end{figure*}

%% file: figure/robust.tex
\begin{figure}[!t]
      \centering
      \includegraphics[scale=0.55]{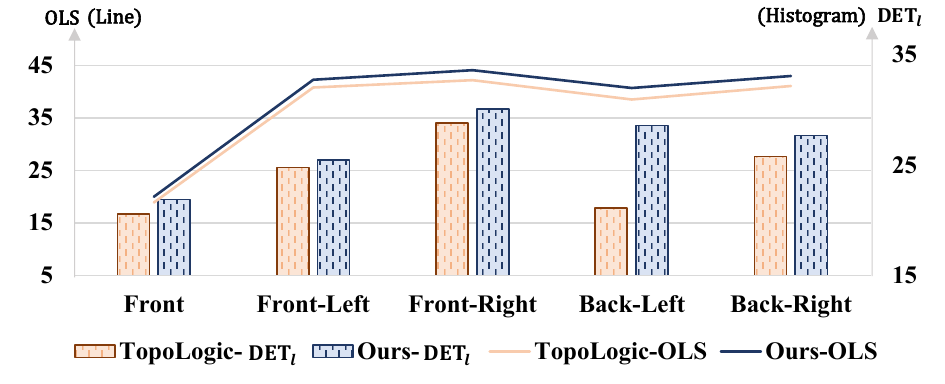}
      \caption{Comparative results under missing-view conditions.
      Across multiple scenarios with missing views, the proposed method achieves better overall performance than TopoLogic and also shows advantages in instance detection accuracy.
      }
      \label{Fig:robust}  
\end{figure}

%% file: table/noise_ex.tex
\begin{table}[!t]
\fontsize{10}{13}\selectfont
\renewcommand{\arraystretch}{1.1}
\setlength\tabcolsep{9pt}
\caption{Robustness analysis under environmental changes.}
\label{tab:noise}
\begin{center}
\resizebox{1.0\linewidth}{!}{\begin{tabular}{l|cc||c|cc}
\hline 
 \multirow{2}{*}{Method }  &\multicolumn{2}{c||}{OLS}  & \multirow{2}{*}{Noise} &\multirow{2}{*}{$\text{DET}_{\textit{l}}$} &\multirow{2}{*}{OLS} \\ \cline{2-3}
 & (0-$\pm$30m)  &  ($\pm$30m-$\pm$50m)    &  & & \\ \hline\hline
TopoLogic~\cite{topologic}  &  43.4   &38.9   &Angle \ ($\pm 2 \degree$) &31.2 &45.3 \\
Ours   & 44.0  &39.8 
  &Height \ ($\pm$0.2m)  &31.2  &45.4\\ \hline
\end{tabular}}
\end{center}
\end{table}

%% file: figure/night_show.tex
\begin{figure}[!t]
      \centering
      \includegraphics[scale=0.50]{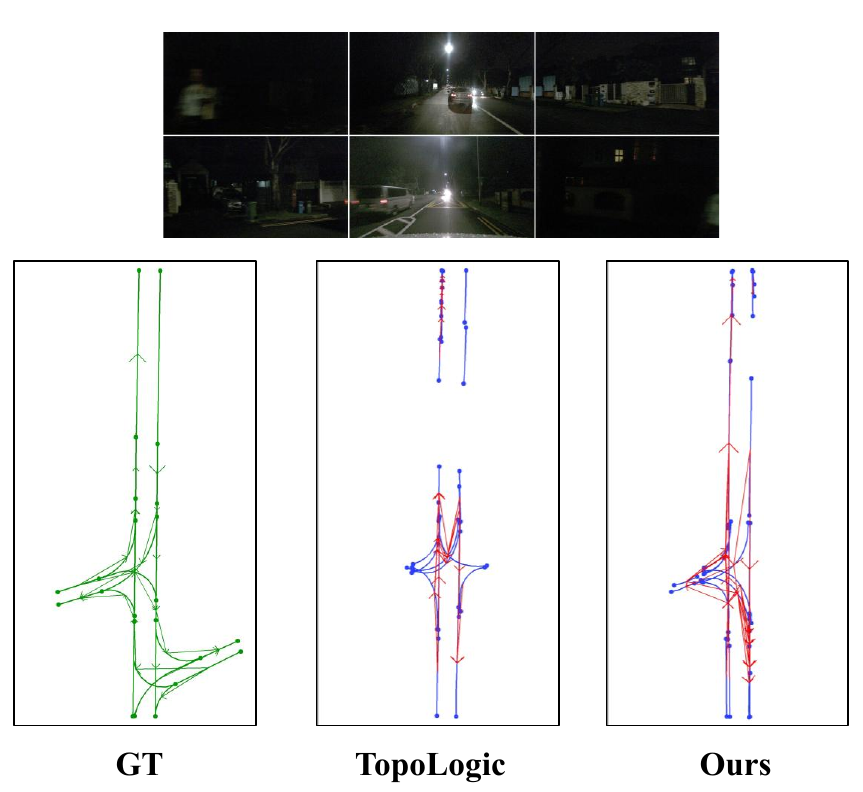}
      \caption{Visualization results in nighttime scenarios. The proposed method detects instances better than TopoLogic~\cite{topologic}, though an accuracy gap to real-world road conditions remains.
      }
      \label{fig:night}  
\end{figure}

%% file: contents/conclution.tex
In this paper, we have proposed HGeo-TopoMap, a hierarchical geometric prior learning framework for topological map construction. Specifically, Geometric Adaptive Learning (GAL) incorporates explicit priors of road structures through spatially adaptive attention, while Geometric Consistency Learning (GCL) integrates center-guided hierarchical decoding with geometry-aware contrastive supervision to enhance centerline representations. By jointly exploiting road-structure guidance and intrinsic geometric regularities, the proposed framework improves centerline detection and topology reasoning. Extensive experiments on the OpenLane-V2 centerline and lane segment benchmarks demonstrate consistent improvements in topological mapping accuracy, while robustness evaluations further verify its effectiveness under challenging driving conditions.

Despite these improvements, the priors of road structures derived from single-frame perspective images remain susceptible to occlusions and incomplete observations, potentially limiting their reliability in complex scenarios. Future work will explore temporal information aggregation to construct more complete and consistent road-structure priors, further improving the robustness of topological map construction.

%% file: main.bbl
\begin{thebibliography}{10}
\providecommand{\url}[1]{#1}
\csname url@samestyle\endcsname
\providecommand{\newblock}{\relax}
\providecommand{\bibinfo}[2]{#2}
\providecommand{\BIBentrySTDinterwordspacing}{\spaceskip=0pt\relax}
\providecommand{\BIBentryALTinterwordstretchfactor}{4}
\providecommand{\BIBentryALTinterwordspacing}{\spaceskip=\fontdimen2\font plus
\BIBentryALTinterwordstretchfactor\fontdimen3\font minus \fontdimen4\font\relax}
\providecommand{\BIBforeignlanguage}[2]{{%
\expandafter\ifx\csname l@#1\endcsname\relax
\typeout{** WARNING: IEEEtran.bst: No hyphenation pattern has been}%
\typeout{** loaded for the language `#1'. Using the pattern for}%
\typeout{** the default language instead.}%
\else
\language=\csname l@#1\endcsname
\fi
#2}}
\providecommand{\BIBdecl}{\relax}
\BIBdecl

\bibitem{hdmapreview}
I.~Abuhadrous \emph{et~al.}, ``Vectorized {HD} map learning for autonomous driving: A survey of representations, architectures, and future directions,'' \emph{IEEE Transactions on Intelligent Transportation Systems}, vol.~27, no.~10, pp. 11\,929--11\,948, 2026.

\bibitem{mapqp}
R.~Li \emph{et~al.}, ``{MapQP:} {Relational} query and map-aware position encoding for online {HD} map construction,'' \emph{IEEE Transactions on Intelligent Transportation Systems}, vol.~27, no.~10, pp. 12\,233--12\,246, 2026.

\bibitem{mo2026mapuq}
S.~Mo \emph{et~al.}, ``{MapUQ:} {Map} with uncertainty quantification for robust {BEV} vectorized construction,'' in \emph{Proc. ICML}, 2026, pp. 89\,561--89\,578.

\bibitem{zhang2025local}
Y.~Zhang \emph{et~al.}, ``Local vectorized high definition map construction for autonomous driving: A comprehensive review,'' \emph{IEEE Transactions on Intelligent Transportation Systems}, vol.~26, no.~12, pp. 21\,502--21\,525, 2025.

\bibitem{hdmapnet}
Q.~Li \emph{et~al.}, ``{HDMapNet:} {An} online {HD} map construction and evaluation framework,'' in \emph{Proc. ICRA}, 2022, pp. 4628--4634.

\bibitem{maptrv2}
B.~Liao \emph{et~al.}, ``{MapTRv2:} {An} end-to-end framework for online vectorized {HD} map construction,'' \emph{International Journal of Computer Vision}, vol. 133, p. 1352–1374, 2025.

\bibitem{review}
Y.~Yao \emph{et~al.}, ``A concise survey on lane topology reasoning for {HD} mapping,'' in \emph{Proc. IV}, 2025, pp. 468--475.

\bibitem{mappath}
H.~Wang \emph{et~al.}, ``Toward trustworthy autonomous driving through the quantification and propagation of uncertainty from {HD} maps to trajectory prediction,'' \emph{IEEE Transactions on Intelligent Transportation Systems}, vol.~27, no.~11, pp. 14\,106--14\,117, 2026.

\bibitem{sun2026diffsemanticfusion}
Z.~Sun \emph{et~al.}, ``{DiffSemanticFusion:} {Semantic} raster {BEV} fusion for autonomous driving via online map diffusion,'' \emph{IEEE Robotics and Automation Letters}, vol.~11, no.~3, pp. 2354--2361, 2026.

\bibitem{toponet}
T.~Li \emph{et~al.}, ``Graph-based topology reasoning for driving scenes,'' \emph{Science China Information Sciences}, vol.~69, no.~5, p. 152103, 2026.

\bibitem{topologic}
Y.~Fu \emph{et~al.}, ``{TopoLogic:} {An} interpretable pipeline for lane topology reasoning on driving scenes,'' in \emph{Proc. NeurIPS}, 2024, pp. 61\,658--61\,676.

\bibitem{lanesegnet}
T.~Li \emph{et~al.}, ``{LaneSegNet:} {Map} learning with lane segment perception for autonomous driving,'' in \emph{Proc. ICLR}, 2024, pp. 1708--1724.

\bibitem{mask2former}
B.~Cheng \emph{et~al.}, ``Masked-attention mask transformer for universal image segmentation,'' in \emph{Proc. CVPR}, 2022, pp. 1280--1289.

\bibitem{segformer}
E.~Xie \emph{et~al.}, ``{SegFormer:} {Simple} and efficient design for semantic segmentation with transformers,'' in \emph{Proc. NeurIPS}, 2021, pp. 12\,077--12\,090.

\bibitem{genmapping}
S.~Li \emph{et~al.}, ``{GenMapping:} {Unleashing} the potential of inverse perspective mapping for robust online {HD} map construction,'' \emph{IEEE Open Journal of Intelligent Transportation Systems}, vol.~7, pp. 1492--1506, 2026.

\bibitem{wang2023openlanev2}
H.~Wang \emph{et~al.}, ``{OpenLane-V2:} {A} topology reasoning benchmark for unified {3D} {HD} mapping,'' in \emph{Proc. NeurIPS}, 2023, pp. 18\,873--18\,884.

\bibitem{LSS}
J.~Philion and S.~Fidler, ``Lift, splat, shoot: Encoding images from arbitrary camera rigs by implicitly unprojecting to {3D},'' in \emph{Proc. ECCV}, 2020, pp. 194--210.

\bibitem{bevdepth}
Y.~Li \emph{et~al.}, ``{BEVDepth:} {Acquisition} of reliable depth for multi-view {3D} object detection,'' in \emph{Proc. AAAI}, 2023, pp. 1477--1485.

\bibitem{li2022bevformer}
Z.~Li \emph{et~al.}, ``{BEVFormer:} {Learning} bird's-eye-view representation from multi-camera images via spatiotemporal transformers,'' in \emph{Proc. ECCV}, 2022, pp. 1--18.

\bibitem{gkt}
H.~Li \emph{et~al.}, ``Efficient and robust {2D-to-BEV} representation learning via geometry-guided kernel transformer,'' \emph{arXiv:2206.04584}, 2022.

\bibitem{PanopticBEV}
N.~Gosala and A.~Valada, ``Bird’s-eye-view panoptic segmentation using monocular frontal view images,'' \emph{IEEE Robotics and Automation Letters}, vol.~7, no.~2, pp. 1968--1975, 2022.

\bibitem{zhu2020deformable}
X.~Zhu \emph{et~al.}, ``Deformable {DETR}: {Deformable} transformers for end-to-end object detection,'' in \emph{Proc. ICLR}, 2021.

\bibitem{yolo}
J.~Redmon \emph{et~al.}, ``You only look once: Unified, real-time object detection,'' in \emph{Proc. CVPR}, 2016, pp. 779--788.

\bibitem{liao2022maptr}
B.~Liao \emph{et~al.}, ``{MapTR:} {Structured} modeling and learning for online vectorized {HD} map construction,'' in \emph{Proc. ICLR}, 2023.

\bibitem{Yuan_2024_streammapnet}
T.~Yuan \emph{et~al.}, ``{StreamMapNet:} {Streaming} mapping network for vectorized online {HD} map construction,'' in \emph{Proc. WACV}, 2024, pp. 7341--7350.

\bibitem{topo2seq}
Y.~Yang \emph{et~al.}, ``{Topo2Seq:} {Enhanced} topology reasoning via topology sequence learning,'' in \emph{Proc. AAAI}, 2025, pp. 9318--9326.

\bibitem{kalfaoglu2025topobda}
M.~E. Kalfaoglu \emph{et~al.}, ``{TopoBDA:} {Towards} bezier deformable attention for road topology understanding,'' \emph{Neurocomputing}, p. 132360, 2025.

\bibitem{topohr}
Y.~Bai \emph{et~al.}, ``{TopoHR:} {Hierarchical} centerline representation for cyclic topology reasoning in driving scenes with point-to-instance relations,'' in \emph{Proc. CVPR}, 2026, pp. 1--10.

\bibitem{li2025reusing}
Y.~Li \emph{et~al.}, ``Reusing attention for one-stage lane topology understanding,'' in \emph{Proc. IROS}, 2025, pp. 16\,977--16\,984.

\bibitem{fu2026topopoint}
Y.~Fu \emph{et~al.}, ``{TopoPoint:} {Enhance} topology reasoning via endpoint detection in autonomous driving,'' in \emph{Proc. NeurIPS}, 2025, pp. 118\,229--118\,250.

\bibitem{unitopo}
H.~Li \emph{et~al.}, ``Unified modeling of lane and lane topology for driving scene reasoning,'' \emph{IEEE Transactions on Circuits and Systems for Video Technology}, vol.~36, no.~9, pp. 12\,637--12\,648, 2026.

\bibitem{xu2026fine_grained}
G.~Xu \emph{et~al.}, ``Fine-grained representation for lane topology reasoning,'' in \emph{Proc. AAAI}, 2026, pp. 11\,214--11\,222.

\bibitem{topomlp}
D.~Wu \emph{et~al.}, ``{TopoMLP:} {A} simple yet strong pipeline for driving topology reasoning,'' in \emph{Proc. ICLR}, 2024, pp. 45\,604--45\,615.

\bibitem{li2025ratopo}
H.~Li \emph{et~al.}, ``{RATopo:} {Improving} lane topology reasoning via redundancy assignment,'' in \emph{Proc. MM}, 2025, pp. 777--786.

\bibitem{topoformer}
C.~Lv \emph{et~al.}, ``{T2SG:} {Traffic} topology scene graph for topology reasoning in autonomous driving,'' in \emph{Proc. CVPR}, 2025, pp. 17\,197--17\,206.

\bibitem{realtopo}
Y.~Luo \emph{et~al.}, ``{RelTopo:} {Multi-level} relational modeling for driving scene topology reasoning,'' \emph{arXiv:2506.13553}, 2025.

\bibitem{kalfaoglu2026topomaskv3}
M.~E. Kalfaoglu \emph{et~al.}, ``{TopoMaskV3:} {3D} mask head with dense offset and height predictions for road topology understanding,'' in \emph{Proc. CVPR}, 2026, pp. 675--684.

\bibitem{liang2026persistent}
S.~Liang \emph{et~al.}, ``Persistent autoregressive mapping with traffic rules for autonomous driving,'' in \emph{Proc. AAAI}, 2026, pp. 6862--6870.

\bibitem{pmapnet}
Z.~Jiang \emph{et~al.}, ``{P-MapNet:} {Far-seeing} map generator enhanced by both {SDMap} and {HDMap} priors,'' \emph{IEEE Robotics and Automation Letters}, vol.~9, no.~10, pp. 8539--8546, 2024.

\bibitem{SMART}
J.~Ye \emph{et~al.}, ``{SMART:} {Advancing} scalable map priors for driving topology reasoning,'' in \emph{Proc. ICRA}, 2025, pp. 3298--3304.

\bibitem{pham2025coherent}
K.~S. Pham \emph{et~al.}, ``Coherent online road topology estimation and reasoning with standard-definition maps,'' in \emph{Proc. IROS}, 2025, pp. 9886--9893.

\bibitem{monninger2026navmapfusion}
T.~Monninger \emph{et~al.}, ``{NavMapFusion:} {Diffusion-based} fusion of navigation maps for online vectorized {HD} map construction,'' in \emph{Proc. WACV}, 2026, pp. 7945--7954.

\bibitem{priorglobal}
S.~M. Bateman \emph{et~al.}, ``Exploring real world map change generalization of prior-informed {HD} map prediction models,'' in \emph{Proc. CVPRW}, 2024, pp. 4568--4578.

\bibitem{globalmapnet}
A.~Shi \emph{et~al.}, ``{GlobalMapNet:} {An} online framework for vectorized global {HD} map construction,'' \emph{arXiv:2409.10063}, 2024.

\bibitem{li2026amap}
R.~Li \emph{et~al.}, ``{AMap:} {Distilling} future priors for ahead-aware online {HD} map construction,'' in \emph{Proc. CVPR}, 2026, pp. 24\,906--24\,917.

\bibitem{shon2026_nonvisible}
S.~Shon \emph{et~al.}, ``{D2HDMap:} {Non-visible} driveline map prior for online vectorized {HD} map prediction,'' in \emph{Proc. IV}, 2026, pp. 1156--1162.

\bibitem{lengerer2026aerialfusionmapnet}
D.~Lengerer \emph{et~al.}, ``{AerialFusionMapNet:} {Online} {HD} map construction with aerial-onboard {BEV} fusion,'' in \emph{Proc. ITSC}, 2026.

\bibitem{immel2025sdtagnet}
F.~Immel \emph{et~al.}, ``{SDTagNet:} {Leveraging} text-annotated navigation maps for online {HD} map construction,'' in \emph{Proc. NeurIPS}, 2025, pp. 30\,175--30\,201.

\bibitem{chen2026vlms}
X.~Chen \emph{et~al.}, ``Are {VLMS} ready for lane topology awareness in autonomous driving?'' in \emph{Proc. ICASSP}, 2026, pp. 19\,787--19\,791.

\bibitem{zhang2026dihmap}
G.~Zhang \emph{et~al.}, ``{DiHMap:} {Direction-consistent} hybrid learning for vectorized {HD} map construction,'' \emph{IEEE Transactions on Intelligent Transportation Systems}, 2026.

\bibitem{mgmap}
X.~Liu \emph{et~al.}, ``{MGMap:} {Mask-guided} learning for online vectorized {HD} map construction,'' in \emph{Proc. CVPR}, 2024, pp. 14\,812--14\,821.

\bibitem{priorvecmap}
C.~Zhang \emph{et~al.}, ``Improving hierarchical representations of vectorized {HD} maps with perspective clues,'' \emph{IEEE Robotics and Automation Letters}, vol.~10, no.~12, pp. 12\,533--12\,540, 2025.

\bibitem{topo2d}
H.~Li \emph{et~al.}, ``Enhancing {3D} lane detection and topology reasoning with {2D} lane priors,'' \emph{arXiv:2406.03105}, 2024.

\bibitem{uncer}
K.~Sirohi \emph{et~al.}, ``Uncertainty-aware panoptic segmentation,'' \emph{IEEE Robotics and Automation Letters}, vol.~8, no.~5, pp. 2629--2636, 2023.

\bibitem{Mapillary}
G.~Neuhold \emph{et~al.}, ``The mapillary vistas dataset for semantic understanding of street scenes,'' in \emph{Proc. ICCV}, 2017, pp. 4990--4999.

\end{thebibliography}
